\documentclass{article} 
\usepackage{iclr2021_conference,times}
\usepackage{graphicx}
\usepackage{multicol}

\usepackage{amsmath,amsfonts,bm}

\def\eqref#1{equation~\ref{#1}}

\def\1{\bm{1}}

\DeclareMathAlphabet{\mathsfit}{\encodingdefault}{\sfdefault}{m}{sl}
\SetMathAlphabet{\mathsfit}{bold}{\encodingdefault}{\sfdefault}{bx}{n}

\usepackage{hyperref}
\usepackage{url}

\title{\textit{EfficientTempNet: Temporal Super-Resolution of Radar Rainfall}}

\author{%
  Bekir Z Demiray \\
  IIHR — Hydroscience \& Engineering\\
  University of Iowa\\
  \texttt{bekirzahit-demiray@uiowa.edu}
  \And
  Muhammed Sit \\
  IIHR — Hydroscience \& Engineering\\
  University of Iowa\\
  \texttt{muhammed-sit@uiowa.edu}
  \AND
  Ibrahim Demir \\
  IIHR — Hydroscience \& Engineering\\
  University of Iowa\\
  \texttt{ibrahim-demir@uiowa.edu}
}

\begin{document}

\maketitle

\begin{abstract}
Rainfall data collected by various remote sensing instruments such as radars or satellites has different space-time resolutions. This study aims to improve the temporal resolution of radar rainfall products to help with more accurate climate change modeling and studies. In this direction, we introduce a solution based on EfficientNetV2, namely EfficientTempNet, to increase the temporal resolution of radar-based rainfall products from 10 minutes to 5 minutes. We tested EfficientRainNet over a dataset for the state of Iowa, US, and compared its performance to three different baselines to show that EfficientTempNet presents a viable option for better climate change monitoring.
\end{abstract}

\section{Introduction}

The importance of environmental data has grown significantly in recent years due to the increased impact of natural disasters. Rainfall data, in particular, plays a crucial role in various climate modeling applications, such as flood forecasting  \cite{cite1, cite2}, monitoring water quality  \cite{cite3}, or managing wastewater \cite{cite4}. Given that the spatial and temporal patterns of rain are crucial in these modeling efforts, having reliable and accessible precipitation maps is vital for advancing research on climate related hazards with different objectives like risk assessment \cite{cite5} or disaster mitigation \cite{cite6}.
\\
\\
Quantitative Precipitation Estimation (QPE) systems provide rainfall data that takes into account three dimensions, which include latitude and longitude as the spatial coordinates and temporal resolution as the third dimension. Weather radars are the primary source used in QPE, and they allow to record the space-time characteristics of precipitation, which is essential for making accurate streamflow predictions in hydrology. Improving rainfall datasets in radar hydrology largely involves addressing uncertainty factors, while the focus is on acquiring more precise precipitation data to enhance our understanding of weather patterns in terms of space and time. However, once the data is obtained, the task of creating better datasets becomes a separate challenge.
\\

The temporal resolution of rainfall data is a critical factor in determining the accuracy of predictive modeling efforts (e.g., \cite{cite11}). This paper aims to address the issue of low temporal resolution rainfall products by proposing a convolutional neural network to enhance the temporal resolution of rainfall data. The proposed CNN model, EfficientTempNet, is based on EfficientNetV2 \cite{cite12}, and the performance of the network is compared to three different methods: the nearest frame, optical flow, and TempNet \cite{cite13}.


\subsection{Related Work}
Rainfall products aren’t the only data that is subject to temporal interpolation between two 2D maps. Various studies in computer vision literature were presented employing neural networks based approaches for video frame interpolation \cite{cite14_5, cite15_6, cite16_7, cite17_8}. Conversely, the literature about how to interpolate time in rainfall datasets is limited. In \cite{cite18_9}, researchers used advection correction to create 1-minute rainfall maps from 5-minute ones.  Building upon \cite{cite18_9}, in \cite{cite13}, a residual but simple CNN architecture, namely TempNet, was proposed and was shown to be superior to the dense optical flow-based advection correction method and a non-residual CNN over the IowaRain \cite{cite19_10} dataset. To the best of our knowledge, TempNet is the only study that tackles the problem of temporal super-resolution of radar rainfall products using neural networks, thus forming the baseline of this study. Consequently, in a similar fashion, in this study, the performance of EfficientTempNet will be presented over the IowaRain dataset to show the increment in performance.
\\
\\
The paper is organized as follows: Section 2 introduces the dataset used and provides an overview of the methodology. Section 3 presents and discusses the preliminary results of the comparison between all methods. Finally, Section 4 summarizes the findings.

\section{Methods}
\subsection{Data}
\begin{figure}[ht]
\includegraphics[height=4cm]{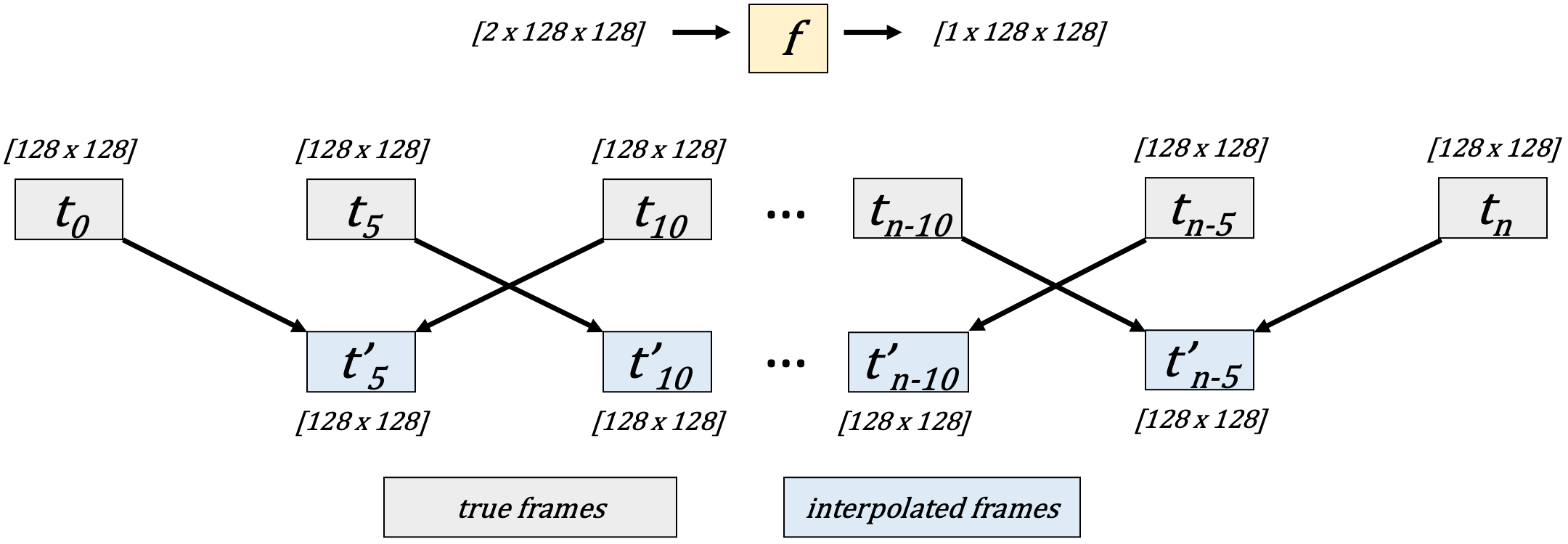}
  \centering
  \caption{Input/Output shapes and problem definition for EfficientTempNet}
  \label{fig-data}
\end{figure}

IowaRain \cite{cite19_10, cite31_additional} is a rainfall event dataset that covers the years of 2016 and 2019 with 5-minute temporal resolution. The dataset covers an area that bounds the state of Iowa and has the size of 1088x1760 pixels with 500m of spatial resolution. In order to meet memory bottlenecks and computational complexity drawbacks, we sampled an area within the IowaRain domain from eastern Iowa that is 768x768 in size; then we averaged the values in the area to downscale the rainmap spatial resolution to 3 km, which effectively changed the rainfall map sizes to 128x128. After the subsampling, the event detection criteria of IowaRain were applied to the new area; 82, 83, 90, and 110 events were obtained for each year in 2016 and 2019, respectively. To get a more approximate 70/30 split when dividing the dataset, we used the rainfall events in 2019 as the test set and all the rainfall events before 2019 as the train set, for total set sizes of 255 and 110 events. For each snapshot, or 2D map, $t_{s}$ from a rain event, a snapshot immediately following it $t_{s+5}$ and immediately preceding it $t_{s-5}$ was converted into its own entry in the dataset. In the end, for each dataset entry or sample, there were three 2D rainfall maps, for $t_{s-5}$, $t_{s+5}$, and $t_{s}$, the first two being inputs and the last one being the output (Figure \ref{fig-data}). In the end, there were 19,264 train and 7,762 test entries used in this study.

\subsection{EfficientTempNet}
\begin{figure}
\includegraphics[width=\textwidth]{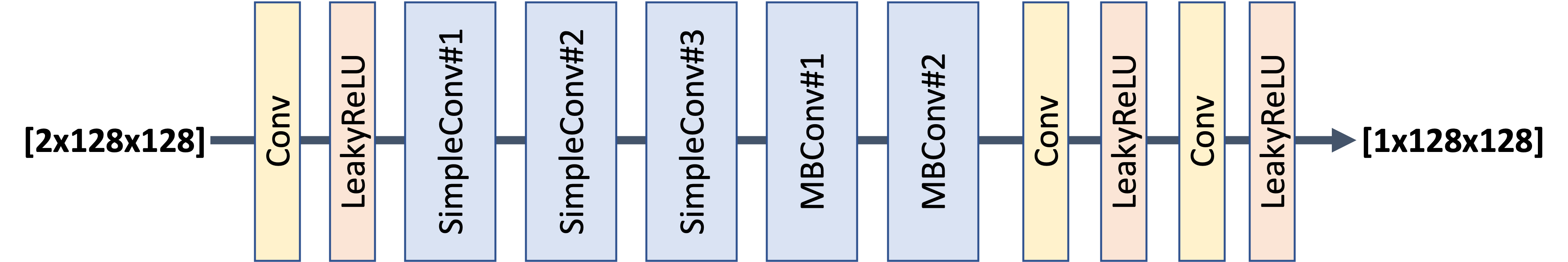}
  \centering
  \caption{Architecture of Proposed Method}
  \label{fig-network}
\end{figure}

The foundation of this work is based on EfficientNetV2 \cite{cite12}. In order to create a model that takes two rainfall maps and outputs another one with the same size, EfficientNetV2 is altered by modifying its key component, MBConv block \cite{cite20}, which was not specifically developed with this task in mind. Our model takes two 2D rain map and combines them prior to passing them into a convolutional layer with 24 feature maps. After this layer, multiple SimpleConv and MBConv blocks are used to extract information. In the last section of the model, two convolutional layers help to get the desired output. Our model is depicted visually in Figure \ref{fig-network}.
\\
\\
In EfficientNetV2, there are two different blocks, namely MBConv and Fused-MBConv. In the proposed method, we modify the MBConv blocks, and instead of Fused-MBConv blocks, SimpleConv blocks were used, as our experiments during model development favored them over Fused-MBConv blocks. In our MBConv blocks, batch normalization layers are removed, and activation layers are changed with LeakyReLU. The remaining parts of the MBConv are the same as the original implementation, including the SE layer and depthwise convolutions. In our SimpleConv blocks, two convolutional layers with kernel sizes 3 and 1 are used with LeakyReLU in between them. Similar to the MBConv block, SimpleConv blocks take advantage of residual connections. In addition to these, input size doesn’t change throughout the model. Details of SimpleConv and MBConv blocks are provided in Table \ref{table1}.
\begin{table}[ht]
\caption{Blocks' details in our model}
\label{table1}
\centering
\resizebox{\textwidth}{!}{%
\begin{tabular}{|c|c|c|c|c|c|}
\hline
                & \textbf{SimpleConv\#1} & \textbf{SimpleConv\#2} & \textbf{SimpleConv\#3} & \textbf{MBConv\#1} & \textbf{MBConv\#2} \\ \hline
\textbf{\# of Channels}  & 24            & 48            & 64            & 128       & 256       \\ \hline
\textbf{Expansion ratio} & 1             & 4             & 4             & 4         & 6         \\ \hline
\textbf{\# of Layers}    & 2             & 4             & 3             & 4         & 8         \\ \hline
\end{tabular}%
}
\end{table}
\\
\\
The RAdam optimizer \cite{cite21} was employed for training with a learning rate of 0.001, and the Mean Absolute Error as the loss function and for evaluation. The implementation of the network was done using Pytorch 1.9 \cite{cite22}, and the ReduceLROnPlateau scheduler was utilized to adjust the learning rate downward if there was no progress in reducing the model’s loss over three consecutive epochs. The training was performed using NVIDIA Titan V GPUs.

\section{Results}
This section outlines the metrics used to evaluate performance and presents the results of our methods as well as three baselines: nearest frame, optical flow, and TempNet. First, we will give information about compared methods, then describe the metrics, and finalize the section with scores and discussion.
\\
\\
\textbf{Nearest Frame} - The nearest frame involves assuming that the interpolated frame is equal to the closest frame in time to the forecasted frame. In our case, we decided to select the predecessor frame as the closest, although both frames are in the same proximity.
\\
\\
\textbf{Optical Flow} - Although a variety of optical flow calculation algorithms can be found in the computer vision literature, the Gunnar-Farneback optical flow was used in this paper \cite{cite24}. Each value’s pixel intensity was determined using the Gunnar-Farneback optical flow. For the rain map scenario, this would entail figuring out the shifts in each measurement over the course of the two-dimensional rain map’s 3km by 3km grid. Once the optical flow is computed, all measurements are shifted between frames based on their position in the first frame and their motion vectors in the optical flow. NumPy \cite{cite25}, a library for numerical computation, and OpenCV \cite{cite26}, a library for computer vision, were utilized in implementation.
\\
\\
\textbf{TempNet} - TempNet has done a similar job to ours. The study is powered by three components, and the design is completed with residual connections. More details on the TempNet can be found here \cite{cite13}.
\\
\\
We measured the performance of each of the previously mentioned methods using four metrics, namely, Mean Absolute Error (MAE), Probability of Detection (POD), and False Alarm Ratio (FAR) and Critical Success Index (CSI). The MAE calculates the average of the absolute differences between the estimated and actual 2D rain maps in the test dataset. The POD \ref{POD}, FAR \ref{FAR}, and CSI \ref{CSI} metrics are calculated using the number of hits (H), false alarms (F), and misses (M), respectively, in a binary manner. H represents the number of correctly estimated rainfall cells, meaning the number of elements in the 2D map that were correctly estimated as non-zero values. F represents the number of wrongly estimated rainfall cells, where the cells were estimated to have rain, but the corresponding indices in the ground truth 2D map had zero. M represents the number of rainfall cells that were estimated as zero but had non-zero values in the ground truth. It’s important to note that a value of 1.0 is best for POD and CSI, whereas a value of 0.0 is best for FAR. All metrics were calculated using a threshold value of 0.0001 over the estimated rainfall maps, since the neural networks would produce small non-zero values throughout the estimated 2D rainfall maps.
\begin{multicols}{3}
  \begin{equation}
    POD=\frac{H}{H+M}
    \label{POD}
  \end{equation}\break
  \begin{equation}
    FAR=\frac{F}{H+F}
    \label{FAR}
  \end{equation}\break
   \begin{equation}
    CSI = \frac{H}{H+F+M}
    \label{CSI}
  \end{equation} 
\end{multicols}
\begin{table}[ht]
\caption{Performance summary of tested methods for predicting intermediate frame}
\label{table2}
\centering
\begin{tabular}{|c|c|c|c|c|}
\hline
\textbf{Methodology}      & \textbf{MAE(mm/h)↓} & \textbf{CSI ↑} & \textbf{POD ↑} & \textbf{FAR ↓} \\ \hline
\textbf{Nearest Frame}    & 0.630               & 0.840          & 0.911          & 0.0878         \\ \hline
\textbf{Optical Flow}     & 0.324               & 0.857          & \textbf{0.978}          & 0.127          \\ \hline
\textbf{TempNet}          & 0.272               & 0.898          & 0.919          & \textbf{0.0252}         \\ \hline
\textbf{EfficientTempNet} & \textbf{0.208}               & \textbf{0.923}          & 0.972          & 0.0524         \\ \hline
\end{tabular}
\end{table}
Table \ref{table2} presents the performance of the baselines as well as the CNN-based models on the test set in terms of described metrics for the interpolation of the frame at $t_{s}$ from frames at $t_{s-5}$ and $t_{s+5}$. As shown in Table \ref{table2}, both TempNet and EfficientTempNet outperform the baseline methods in terms of the MAE metric, which was used to train them. Between EfficientTempNet and TempNet, EfficientTempNet shows more strong results with a significant margin in MAE scores as a result of an increase in model size. As for the POD, optical flow provides the best result with a small margin in comparison to EfficientTempNet. However, with consideration of FAR and POD together, it can be argued that optical flow tends to generate non-zero values that results in correctly identifying true positives but also causes a higher number of false positives compared to EfficientTempNet. Overall, in improving temporal resolution of radar rainfall products, the EfficientTempNet offers a better solution compared to other employed methods since its scores are either best or runner-up over the used metrics.

\section{Conclusion}
In this study, the EfficientNetV2-based CNN model, EfficientTempNet, was introduced for improving the temporal resolution of radar rainfall products and compared to three additional methods. The results showed that EfficientTempNet outperformed the other approaches in terms of various performance metrics, including MAE. This work represents significant progress in creating improved rainfall maps for various purposes, such as flood forecasting \cite{cite27, cite28} and climate change modeling \cite{cite29, cite30}.

\bibliography{iclr2021_conference}
\bibliographystyle{iclr2021_conference}

\end{document}